\documentclass[letterpaper, 10 pt, conference]{ieeeconf}  

\IEEEoverridecommandlockouts                              

\usepackage{graphicx}
\usepackage{subcaption}
\usepackage{booktabs}
\usepackage{multirow}
\usepackage{multicol}
\usepackage{pgfplots}
\usepackage[breakable, skins]{tcolorbox}
\usepackage{amsmath}
\usepackage{tabularray}
\usepackage{algorithmic}
\usepackage{algorithm}
\usepackage{float}
\usepackage{hyperref}
\hypersetup{
    colorlinks=true,
    urlcolor=blue,
    linkcolor=black
}

\newcommand{\static}{$S$}
\newcommand{\none}{$R^-$}
\newcommand{\checkonly}{$R^-_V$}
\newcommand{\ragonly}{$R^+$}
\newcommand{\ragcheck}{$R^+_V$}
\newcommand{\ragsearch}{$R^S$}
\newcommand{\ragsearchcheck}{$R^S_V$}

\title{\LARGE \bf
L3M+P: Lifelong Planning with Large Language Models
}

\author{
Krish Agarwal$^{*\dagger}$, Yuqian Jiang$^{*\dagger}$, Jiaheng Hu$^{\dagger}$, Bo Liu$^{\dagger}$, Peter Stone$^{\dagger\mathsection}$
\thanks{*Equal contribution.}
\thanks{$^\dagger$Department of Computer Science, The University of Texas at Austin \texttt{\{krishagarwal, jiangyuqian, jiahengh\}@utexas,edu, \{bliu, pstone\}@cs.utexas.edu}}
\thanks{$^\mathsection$Sony AI}
}

\begin{document}

\maketitle
\thispagestyle{empty}
\pagestyle{empty}

\begin{abstract}

By combining classical planning methods with large language models (LLMs), recent research such as LLM+P has enabled agents to plan for general tasks given in natural language. However, scaling these methods to general-purpose service robots remains challenging: (1) classical planning algorithms generally require a detailed and consistent specification of the environment, which is not always readily available; and (2) existing frameworks mainly focus on isolated planning tasks, whereas robots are often meant to serve in long-term continuous deployments, and therefore must maintain a dynamic memory of the environment which can be updated with multi-modal inputs and extracted as planning knowledge for future tasks. To address these two issues, this paper introduces L3M+P (Lifelong LLM+P), a framework that uses an external knowledge graph as a representation of the world state. The graph can be updated from multiple sources of information, including sensory input and natural language interactions with humans. L3M+P enforces rules for the expected format of the absolute world state graph to maintain consistency between graph updates. At planning time, given a natural language description of a task, L3M+P retrieves context from the knowledge graph and generates a problem definition for classical planners. Evaluated on household robot simulators and on a real-world service robot, L3M+P achieves significant improvement over baseline methods both on accurately registering natural language state changes and on correctly generating plans, thanks to the knowledge graph retrieval and verification. 

\end{abstract}
\section{INTRODUCTION}
\label{sec:introduction}

Large language models (LLMs) have proven to be very promising natural language (NL) interfaces in a variety of domains, including Robotics \cite{zhang2023llm}. However, their lack of grounding in the physical world prevents them from being used effectively as direct planners in ``agentic" systems \cite{mahowald2024dissociating}. A large body of work has thus been motivated toward grounding LLM-based agents to bridge the gap between human-friendly interfaces and consistent, accurate planning \cite{ahn2022do, liu2023llmp, dagan2023llmdp, yoneda2023statler, song2023llmplanner, bhat2024grounding}.

However, applying these recent advancements to general-purpose service robots remains difficult. A primary challenge is that these robots are meant to serve their function over an extended period of time, and any actions they take must be grounded in a \emph{dynamic}, real-world environment. We can consider the following two cases to see why an LLM alone is insufficient for interfacing with a service robot, and an external structured memory is required.

Suppose a human in the environment informs the service robot on events that have taken place in the environment. These events affect the state of the environment, in turn affecting future planning performed by the service robot. An LLM-only solution might keep track of these events by accumulating them into a report and feeding this report as context to the LLM whenever a planning query is processed. However, the report can grow arbitrarily long, so this approach is prone to model hallucination, which can significantly affect the accuracy of the agent \cite{huang2023hallucination}. As such, the LLM-based agent must have an external memory.

Of course, it is unreasonable to expect a human to report every event that takes place in an environment. Specifically, a service robot can be expected to consume not only human dialogue but also sensor input to gain knowledge about the environment. Solutions already exist for traditional knowledge representation in service robots that can be updated based on sensor input \cite{breux2020knowledge, schmidt-rohr2011knowledge, ICAPS19-Jiang}. In order to use robot perception alongside human dialogue, an LLM must be able to interact with a unified memory that is compatible with a traditional knowledge representation system.

\begin{figure*}
    \centering
    \includegraphics[width=0.8\linewidth]{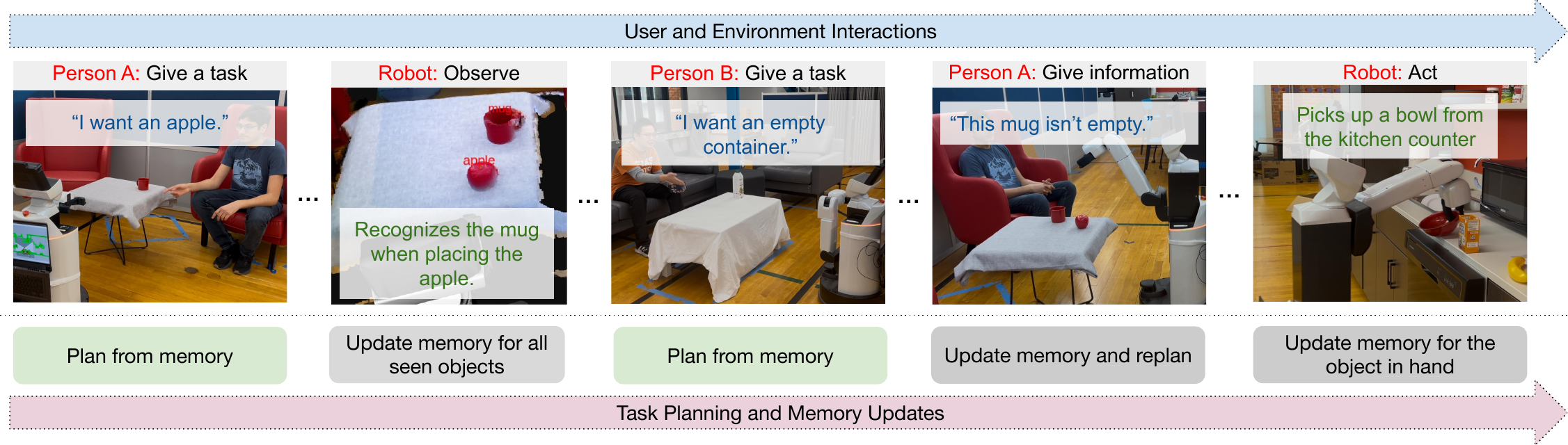}
    \caption{L3M+P enables a household robot to keep a dynamic memory of verbal interactions and other sensory inputs and plan for long-horizon natural language tasks. Top: robot interactions with the environment and users for a sequence of tasks. Bottom: task planning and memory updates. }
    \label{graph:motivating_example}
\end{figure*}

This motivation leads us to develop the L3M+P framework for augmenting existing research on grounded LLM-based agents with a dynamic, structured memory. L3M+P interfaces with this memory as follows.

\begin{enumerate}
    \item It uses a LLM-based natural language interface for updating the memory given NL descriptions of environment updates.
    \item It integrates with robot perception so the memory can be updated based on sensory input.
    \item It retrieves relevant information from the memory that can be used as context within an existing LLM-based, grounded planner.
\end{enumerate}

The rest of this paper is organized as follows: Section~\ref{sec:background} provides preliminary information for the modules in L3M+P. Section~\ref{sec:problem} introduces the formalisms used in this paper. Section~\ref{sec:method} explains the implementation of each component of L3M+P. Section~\ref{sec:experiments} discusses experiments and results. Finally, Section~\ref{sec:related_work} highlights recent work related to L3M+P.
\section{BACKGROUND}
\label{sec:background}

\subsection{Planning with Language Models}

LLM+P \cite{liu2023llmp} is a framework for combining LLMs with classical planning to bridge the gap between NL task descriptions and symbolic planners. These planners operate using the planning domain definition language (PDDL), a commonly-used language to formalize environments and tasks \cite{mcdermott1998pddl, haslum2019introduction}. A PDDL domain defines the state-action space through (a) a set of predicates that can fully represent a state and (b) a set of actions for manipulating the current state. A PDDL problem specifies a set of initial predicates (representing the initial state) and a set of goal conditions (identifying one or more goal states). Given an initial state description, a task description, and a domain, LLM+P prompts an LLM to generate a PDDL problem that a symbolic planner can use along with the domain to generate a plan.

L3M+P extends LLM+P by removing the requirement for a description of the initial state and instead maintaining the current state in a knowledge base that can be used to extract information relevant to a given planning task.

\subsection{Knowledge Graphs}

Knowledge graphs \cite{Ehrlinger2016TowardsAD} are representations that organize information into graph structures, where nodes represent entities and edges represent relationships between entities. A knowledge graph can be constructed from various sources, including structured databases, unstructured text, or other forms of data.

Similar to traditional databases, graph databases also generally support structured querying, such as through the Cypher query language \cite{FrancisCypher}.

The relationship between a pair of nodes in a knowledge graph is commonly referred to as a triplet of subject, predicate, and object. We work with the following triplet forms:
\begin{itemize}
    \item \texttt{(subject, relationship, object)}
    \item \texttt{(subject, property, boolean)}
\end{itemize}

L3M+P uses a knowledge graph to represent the world state for a general-purpose robot.

\subsection{Retrieval-Augmented Generation}

Many applications of LLMs involve answering domain-specific queries. One approach for adapting a pretrained model to this purpose is to fine-tune the model on domain-specific knowledge \cite{wei2022finetune}. However, this method can be expensive and is not versatile to changes in knowledge. Retrieval-augmented generation (RAG) is a method for augmenting language models with external knowledge sources \cite{lewis2020rag}. Context from an external knowledge source is retrieved based on a user-prompt and fed alongside the original prompt to the language model in order to provide the model with sufficient information to answer the prompt.

Our framework has a RAG component which retrieves the relevant edges from a knowledge graph and feeds them as context for the LLM to handle state changes and plan queries.

\begin{figure*}
    \centering
    \begin{subfigure}[b]{0.85\textwidth}
        \includegraphics[page=1,width=1\linewidth]{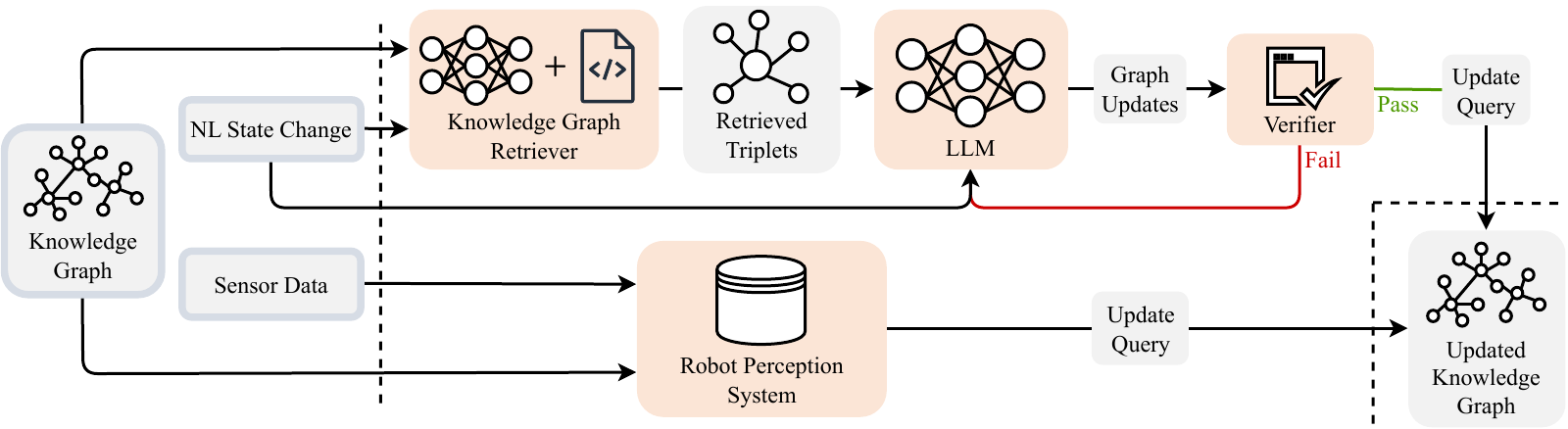}
        \caption{World State Updates}
        \label{diagram:update}
    \end{subfigure}
    \par\medskip
    \begin{subfigure}[b]{0.73\textwidth}
        \includegraphics[page=2,width=1\linewidth]{diagrams.pdf}
        \caption{Plan Generation}
        \label{diagram:plan}
    \end{subfigure}
    \caption{(a) L3M+P keeps its knowledge graph consistent with the real world by receiving sensory input as well as descriptions of changes to the environment from humans. Both sources are used to update the knowledge graph to reflect an up-to-date representation of the environment. Verification is only necessary for Natural Language (NL) updates to guard against LLM hallucinations. (b) L3M+P uses the external knowledge graph to gather relevant context about the state of the environment to integrate with LLM+P for planning without requiring an explicit description of initial state for every planning query.}
    \label{diagram}
\end{figure*}
\section{PROBLEM STATEMENT}
\label{sec:problem}

Our framework aims to solve two distinct but interdependent types of problems: keeping the robot's world state up-to-date, and generating plans based on the current world state.

\subsection{World State Update}

The first problem is to enable the robot to register updates that take place in the environment at any time during its operation. An update is provided either through natural language or through perception, and could represent a single event, multiple events, or even partial knowledge of events.

Formally, the input to an update problem $U$ is the tuple $\langle \mathcal{S}, s_t, u_t \rangle$:
\begin{itemize}
    \item $\mathcal{S}$ is a finite, discrete set of states that represent all possible world states.
    \item $s_t \in \mathcal{S}$ is the robot's knowledge at time $t$. $s_t$ is itself the tuple $\langle V_t, E_t \rangle$, where $V_t$ is a set of entities (vertices) in the environment and $E_t$ is a set of relationships (edges) between the entities at time $t$.
    \item $u_t$ is some verbal/sensory input describing an update taking place at time $t$.
\end{itemize}

\begin{tcolorbox}[
    standard jigsaw,
    title=Example World State Update Problem,
    opacityback=0,
    fontupper=\small,
    fontlower=\small
]
\setlength{\parindent}{15pt}
\noindent
\textbf{State Change:} Gary went to Alexander's bedroom and placed the red pen on the table.\\\\
\textbf{Correct Output:}\\
{\footnotesize\texttt{REMOVE: (red\_pen, in\_person\_hand, gary), (gary, person\_in\_room, jessica\_bedroom)}\\
\texttt{ADD: (gary, person\_in\_room, alexander\_bedroom), (red\_pen, placed\_at\_table, alexander\_bedroom\_table)}}\\\\
\textbf{Explanation:} Gary moved from Jessica's bedroom to Alexander's bedroom. The red pen was placed on the table, leaving Gary's hand.
\end{tcolorbox}

A correct solution to the problem $U$ is the state $s_{t+1} \in \mathcal{S}$ that accurately represents the world state after registering the update $u_t$. If we assume that entities are not added/removed from the environment, the problem can be simplified to only output the change in the relationships $\langle E_t^-, E_t^+ \rangle$, where we interpret $E_{t+1} = E_t - E_t^- + E_t^+$, so $s_{t+1} = \langle V_t, E_{t+1} \rangle$.

\subsection{Plan Generation}

The second problem is to generate a plan to solve a given task. Formally, the input to the planning problem $P$ is the tuple $\langle \mathcal{S}, s_t, g_t, \mathcal{A}, f\rangle$:
\begin{itemize}
    \item $\mathcal{S}$ is again the state space.
    \item $s_t \in \mathcal{S}$ is again the robot's knowledge at time $t$.
    \item $g_t$ is an NL description of a task to be completed by the robot at time $t$.
    \item $\mathcal{A}$ is a set of symbolic actions.
    \item $f: \mathcal{S} \times \mathcal{A} \rightarrow \mathcal{S}$ is the underlying state transition function. $f$ takes the current state and an action and outputs the resulting state.
\end{itemize}

\begin{tcolorbox}[
    standard jigsaw,
    title=Example Plan Generation Problem,
    opacityback=0,
    fontupper=\small,
    fontlower=\small,
]
\setlength{\parindent}{15pt}
\noindent
\textbf{Task:} Turn off the faucet in the bathroom.\\\\
\textbf{Correct Output:}\\
{\footnotesize \texttt{1) (move\_to\_room robot living\_room bathroom)}}\\
{\footnotesize \texttt{2) (turn\_off\_faucet bathroom\_sink bathroom robot)}}\\\\
\textbf{Explanation:} The robot has to move from the living room to the bathroom. It can then turn off the faucet.
\end{tcolorbox}

A solution to the planning problem $P$ is a sequential plan $\pi = \langle a_1, a_2, \dots, a_N \rangle$ with $a_i \in \mathcal{A}$.
\section{METHOD}
\label{sec:method}

We now describe how L3M+P solves the two problems defined above, as shown in Fig.~\ref{diagram}.

\subsection{Knowledge Graph Retrieval}

To represent the state of the environment, L3M+P employs a knowledge graph (KG) that functions as a direct memory base for the agent. Nodes in the graph represent environment entities and edges represent relationships between them. 

KG triplets directly correspond to PDDL predicates. As such, the KG is a full representation of the current state $s \in \mathcal{S}$. Importantly, we assume a PDDL domain that specifies predicates representing the environment state and actions the agent can take is provided. This assumption is reasonable since creating the PDDL domain is a one-time investment, after which the agent acts autonomously over the long-term.

L3M+P retrieves relevant KG nodes and edges as context for update/plan queries. Current RAG pipelines often supply a list of entities in the environment and prompt the LLM to select relevant ones based on a user prompt. Then, from the list of relevant entities, all triplets incoming/outgoing to each entity are extracted at a certain maximum depth. The retrieved triplets serve as context for the downstream task.

When objects are described with attributes, an LLM may struggle to select correct entities by name. We propose a search-based retrieval method (Algorithm~\ref{alg:search_retrieval}) to precisely locate entities matching the description. First, the algorithm prompts the LLM to generate a query graph with entities and their inferred relations from the NL query, and then searches for the most similar grounded KG sub-graph. Algorithm~\ref{alg:subgraph_matching} (DFS) is inspired by recent scene graph approaches supporting single-node queries~\cite{chang2023context,koch2024open3dsg}, employing depth-first search to find entities relevant for state updates or tasks.

\begin{algorithm}[t]
\caption{Search-Based Retrieval Algorithm}\label{alg:search_retrieval}
\begin{algorithmic}[1]
\REQUIRE $V_t, E_t$, entities and relationships in the knowledge graph at timestep $t$
\REQUIRE $u_t$ or $g_t$, the verbal update or plan query
\ENSURE $V_r$, relevant entities
\STATE Prompt the LLM to extract a query graph $<V_q, E_q>$, of ungrounded entities and relationships, from $u_t$ or $g_t$
\STATE Compute the semantic similarity matrix $S$ between $V_q$ and $V_t$
\STATE Sort the entities in $V_q$ by their highest similarity scores
\STATE $M_{optimal} \gets$ {\bf DFS}$(S, V_q, E_q, V_t, E_t, \{\}, \{\})$
\STATE $V_r \gets$ entities mapped by $M_{optimal}$ 
\end{algorithmic}
\end{algorithm}

\begin{algorithm}[t]
\caption{Subgraph Matching Algorithm (DFS)}
\label{alg:subgraph_matching}
\begin{algorithmic}[1]
\REQUIRE $S$, node similarity matrix
\REQUIRE $V_q, E_q, V_t, E_t$
\REQUIRE $M, M_+$, current mapping and the best mapping found
\REQUIRE \mbox{\emph{cutoff}}, a parameter for node similarity cutoffs
\ENSURE $M_{optimal}$, mapping from $V_q$ to $V_t$ with the highest subgraph similarity score
\IF{All nodes in $V_q$ are mapped by $M$}
    \IF{{\bf mapping\_score}$(M) > $ {\bf mapping\_score}$(M_+)$}
        \RETURN $M_{optimal} \gets M$
    \ENDIF
    \RETURN $M_{optimal} \gets M_+$
\ENDIF
\STATE $v_q \gets$ pop front of $V_q$
\STATE Find all entity candidates $v_c \in V_t$ that $S[v_q, v_c] > \mbox{\emph{cutoff}} \times max_v(S[v_q, v])$
\FOR{\textbf{each} candidate $v_c$}
    \STATE Add mapping $v_q \rightarrow v_c$ to $M$
    \STATE ${M_+} \gets$ {\bf DFS}$(S, V_q, E_q, V_t, E_t, M, M_+)$
    \STATE Pop mapping $v_q \rightarrow v_c$ from $M$
\ENDFOR
\RETURN $M_{optimal} \gets M_+$
\end{algorithmic}
\end{algorithm}

\subsection{Updates to Knowledge Graph}

L3M+P dynamically updates the knowledge graph when the agent is provided with external changes in the environment (shown in Fig. \ref{diagram:update}). The updated knowledge graph can enable re-planning a current task right away or solving other tasks (e.g. question answering). The agent can be alerted of these changes in two manners: (1) the robot receives sensory inputs, or (2) a human describes an event in the environment to the agent in natural language. We assume the robot has a perception system that converts observations into a representation (e.g. scene graphs) to update corresponding sub-graphs in the KG~\cite{ICAPS19-Jiang, chang2023context}, and we show an example of such a system in the robot demo. As such, in this section we focus on NL descriptions of updates to the environment.

Any descriptions of environment changes must be reflected in the KG to maintain a consistent memory. Given a description, L3M+P follows Algorithm \ref{alg:kg_update}. In summary, L3M+P

\begin{enumerate}
    \item Retrieves a relevant subgraph from the KG (line \ref{alg:kg_update:retrieve})
    \item Prompts the LLM to generate KG updates given the retrieved subgraph and the update description (line \ref{alg:kg_update:prompt})
    \item Uses the domain PDDL to type check and syntactically verify the generated graph updates (line \ref{alg:kg_update:verify})
    \item Retries if the verifier fails (line \ref{alg:kg_update:retry})
    \item Applies the LLM-generated graph updates when verification succeeds (lines \ref{alg:kg_update:apply1} and \ref{alg:kg_update:apply2})
\end{enumerate}
L3M+P does not insert new entities into the knowledge graph from NL updates. The knowledge graph only tracks the instances of concrete objects seen by the robot.

\begin{algorithm}
\caption{NL Knowledge Graph Updates}
\label{alg:kg_update}
\begin{algorithmic}[1]
\REQUIRE $\langle V_t, E_t \rangle$, the current knowledge graph at time $t$
\REQUIRE $u_t$, the NL update at time $t$
\ENSURE $E_{t+1}$, the updated KG edges (the updated graph is $\langle V_t, E_{t+1} \rangle$)

\STATE $E_t^{rel} \gets \textbf{retriever}(V_t, E_t, u_t)$ \COMMENT{Retrieve a relevant set of edges from the KG} \label{alg:kg_update:retrieve}
\STATE $E_t^{irrel} \gets E_t - E_t^{rel}$
\STATE $prompt \gets [V_t, E_t^{rel}, u_t]$

\REPEAT
    \STATE $output \gets \textbf{LLM}(prompt)$ \COMMENT{Prompt the LLM to generate a KG update} \label{alg:kg_update:prompt}
    \STATE $E_t^+, E_t^- \gets \textbf{parse}(output)$
    \STATE $errors \gets \textbf{verify}(E_t^+, E_t^-, E_t, V_t)$ \label{alg:kg_update:verify}
    \STATE $prompt \gets prompt + [errors]$
\UNTIL{$errors = \emptyset$} \COMMENT{Re-prompt LLM until generated update is valid} \label{alg:kg_update:retry}
\STATE $E_{t+1}^{rel} \gets E_t^{rel} - E_t^- + E_t^+$ \label{alg:kg_update:apply1}
\STATE $E_{t+1} \gets E_t^{irrel} + E_{t+1}^{rel}$ \label{alg:kg_update:apply2}
\RETURN $E_{t+1}$
\end{algorithmic}
\end{algorithm}

\subsection{Planning}

L3M+P uses a modified version of LLM+P to perform planning (shown in Fig. \ref{diagram:plan}). Unlike LLM+P, the user does not provide the full description of the environment. Instead, L3M+P queries the KG to gain sufficient detail for solving a given task (shown in Algorithm \ref{alg:plan}). In summary, L3M+P
\begin{enumerate}
    \item Retrieves a relevant subgraph from the KG (line \ref{alg:plan:retrieve})
    \item Prompts the LLM to generate the \texttt{:goal} block for the PDDL problem given the retrieved subgraph, the PDDL domain, and the task description (line \ref{alg:plan:prompt})
    \item Constructs a PDDL problem with the generated \texttt{:goal} block and using the retrieved subgraph as the \texttt{:init} block (line \ref{alg:plan:problem})
    \item Passes the fully constructed PDDL problem to a symbolic planner alongside the provided PDDL domain to generate a plan (line \ref{alg:plan:planner})
\end{enumerate}

\begin{algorithm}[bt!]
\caption{Planning}
\label{alg:plan}
\begin{algorithmic}[1]
\REQUIRE $\langle V_t, E_t \rangle$, the current knowledge graph at time $t$
\REQUIRE $g_t$, the NL description for a task to complete at time $t$
\REQUIRE $D$, the PDDL domain
\ENSURE $\pi = \langle a_1, a_2, \dots, a_n \rangle$, a sequential plan to accomplish the given task

\STATE $E_t^{rel} \gets \textbf{retriever}(V_t, E_t, g_t)$ \COMMENT{Retrieve a relevant set of edges from the KG} \label{alg:plan:retrieve}
\STATE $goal \gets \textbf{LLM}(V_t, E_t^{rel}, D, g_t)$ \COMMENT{Prompt the LLM to generate a \texttt{:goal} block for the given task} \label{alg:plan:prompt}
\STATE $P \gets \{V_t, E_t^{rel}, goal\}$ \COMMENT{Generate a corresponding PDDL problem ($E_t^{rel}$ populates the \texttt{:init} block)} \label{alg:plan:problem}
\STATE $\pi \gets \textbf{planner}(D, P)$ \label{alg:plan:planner}
\RETURN $\pi$
\end{algorithmic}
\end{algorithm}
\section{EXPERIMENTS}
\label{sec:experiments}

We design experiments to address the following questions:

\begin{enumerate}
    \item Does RAG improve KG update accuracy compared to providing the full KG context to the LLM? That is, does RAG enable smaller prompts to prevent hallucinations, or does the lack of full context worsen performance? \textbf{\textcolor{purple}{(It provides a significant improvement)}}
    \item How much does the verifier improve (or degrade) graph update accuracy? \textbf{\textcolor{purple}{(It gives a decent improvement)}}
    \item Does accurate KG state translate to correct plans, and does incorrect KG state translate to failed plans? \textbf{\textcolor{purple}{(Yes)}}
    \item Does leveraging verbal updates in L3M+P improve the success rates of a service robot in solving tasks? \textbf{\textcolor{purple}{(Yes)}}
\end{enumerate}

To answer these questions, we present three types of experiments. Sec.~\ref{sec:text_based_exp} evaluates how L3M+P updates the KG based on NL state updates as well as generates plans based on a continuously updating KG. This text-based simulation assumes that updates occur at discrete time steps between tasks. Sec.~\ref{sec:ai2_thor_exp} presents the results in an embodied simulator where the plan of the current task must be adapted to both verbal and sensory updates.\footnote{The implementation of the simulation experiments is available at \url{https://github.com/krishagarwal/l3m-p.git}} Sec.~\ref{sec:robot_demo} demonstrates a robot successfully helping users in a home setting where correct KG updates are required to plan current and future tasks.

\subsection{Text-Based Simulation}
\label{sec:text_based_exp}
Since the agent is meant to function as a general-purpose service robot, we introduce an open-ended, text-based household simulator where items are randomly generated, humans randomly manipulate the environment, and random tasks are periodically presented to an agent. The household consists of a diverse set of items (e.g. tables, fridges, sinks, books, food items, dishes, lights, TVs, phones) distributed across various rooms (e.g. kitchen, living room, bedrooms). Humans in the household move objects, interact with faucets/lights, answer phones, etc. The agent is tasked with returning items to certain locations, washing dishes, providing items to humans (like food or drink), etc. Random tasks allow testing the capabilities of the agent in a domain-agnostic environment.

The simulation provides the agent with an NL description of each environment update and an NL description of each planning task. Since this experiment is focused on testing the update and planning capabilities of L3M+P, the simulation directly invokes these functions rather than having the agent infer what function to use. The agent is also provided a domain PDDL file, which, as justified in Section \ref{sec:method}, is reasonable for realistic use cases. To measure accuracy, the simulation maintains a ground-truth KG after each state update and plan execution to compare against L3M+P's proposed KG, and a ground-truth problem PDDL file is also created to generate a ground-truth plan that can be validated against any agent-generated plans. We report success rates for state changes (matching updates in the ground-truth KG) and plans (goal satisfaction in the ground-truth problem).

\begin{table}
\centering
\begin{tblr}{
  cells = {c},
  hlines,
  vlines,
}
Method & State Changes & Plans\\
\static & - & 67.5\\
\none & 71.5 & 72.5\\
\checkonly & 71 & 72.5\\
\ragonly & 77.5 & 80\\
\ragcheck & 92.5 & 85\\
\ragsearch & 96 & 87.5\\
\ragsearchcheck (ours) & \textbf{98} & \textbf{90}
\end{tblr}
\caption{Final state change and plan success rate \% ($\uparrow$) after running the seven agent variants on the simulation.}
\label{sim:results}
\end{table}

We test the following seven variants of the agent against the same simulation:
\begin{enumerate}
    \item \static: The agent receives the full initial KG but does not receive state updates. This is a baseline to measure the sparsity of the instantiated environment/tasks to better compare relative performance of the other ablations.
    \item \none: The agent receives the full KG as context when performing state updates (no RAG).
    \item \checkonly: The agent receives the full KG as state change context (no RAG). State updates are verified against the domain PDDL.
    \item \ragonly: RAG is used to extract relevant context from the KG that fits in the LLM context window to perform state updates.
    \item \ragcheck: RAG is performed as above. State change verification is performed.
    \item \ragsearch: Search-based RAG is performed to extract KG context.
    \item \ragsearchcheck\ (a.k.a L3M+P): Search-based RAG is performed as above. State change verification is performed.
\end{enumerate}

To execute these experiments, we use OpenAI's GPT-4o language model. We also utilize the \textsc{siw-then-bfsf} planner provided by LAPKT \cite{lapkt} as the planner in all experiments.

\begin{tcolorbox}[
    standard jigsaw,
    title=Example of a Failed State Change and a Resulting Incorrect Plan Produced by \none,
    opacityback=0,
    fontupper=\small,
    fontlower=\small,
    breakable
]
\setlength{\parindent}{15pt}
\noindent
\textbf{State Change:} Jessica turned off the overhead light in the laundry room.\\\\
\textbf{Prompt:} \textcolor{purple}{State Change} + \textcolor{purple}{Full KG Context} + Determine which of the relations should be removed and what new relations should be added...\\\\
\textbf{GPT-4o (generated KG update):}\\
\texttt{REMOVE: (laundry\_room\_light, in\_room, laundry\_room)}\\ \texttt{ADD: \emph{empty}}
\tcblower
\textbf{Task:} The water and electricity bills are high. Can you turn off all lights and faucets?\\\\
\textbf{Retrieved KG context (\texttt{:init} block):}\\
\texttt{\ldots}\texttt{(laundry\_room\_light, light\_on, true)}\\\\
\textbf{Prompt:} \textcolor{purple}{Task} + \textcolor{purple}{Full KG Context} + 
Provide the goal block for a problem PDDL file...\\\\
\textbf{GPT-4o (generated \texttt{:goal} block):}\\
\texttt{(:goal (and (forall (?a - light) (not (light\_on ?a))) (forall (?b - sink) (not (faucet\_on ?b)))))}\\\\
\textbf{Plan:}\\
\texttt{
\ldots(turn\_off\_light laundry\_room\_light)
}
\end{tcolorbox}

\begin{figure}[hbt!]
    \centering
    \includegraphics[width=0.8\linewidth]{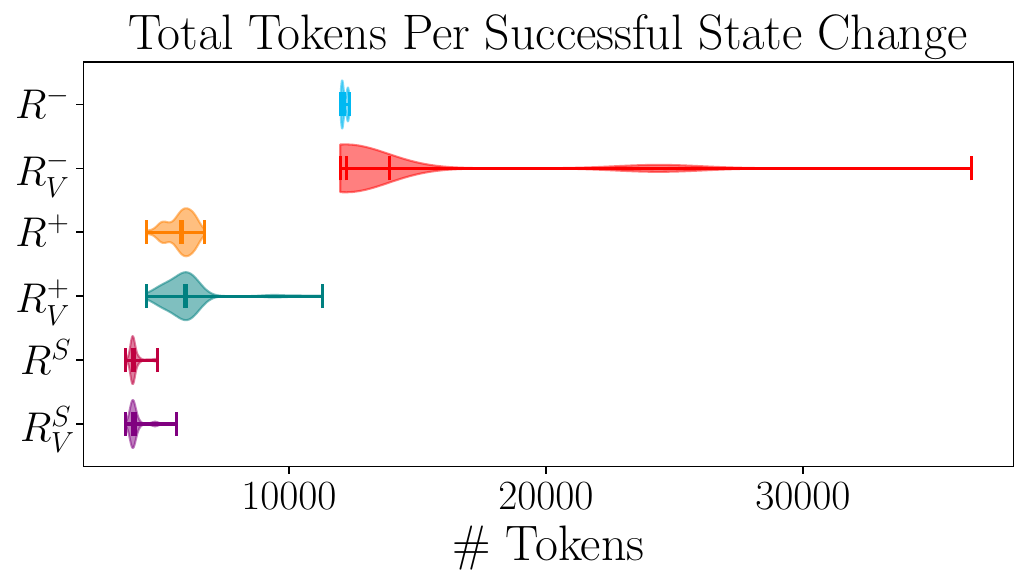}
    \caption{Distributions of number of tokens used per successful state change in the text-based simulation. Variants that use RAG are the most token-efficient.}
    \label{graph:state_change_tokens}
\end{figure}

\begin{figure}[hbt!]
    \centering
    \includegraphics[width=0.8\linewidth]{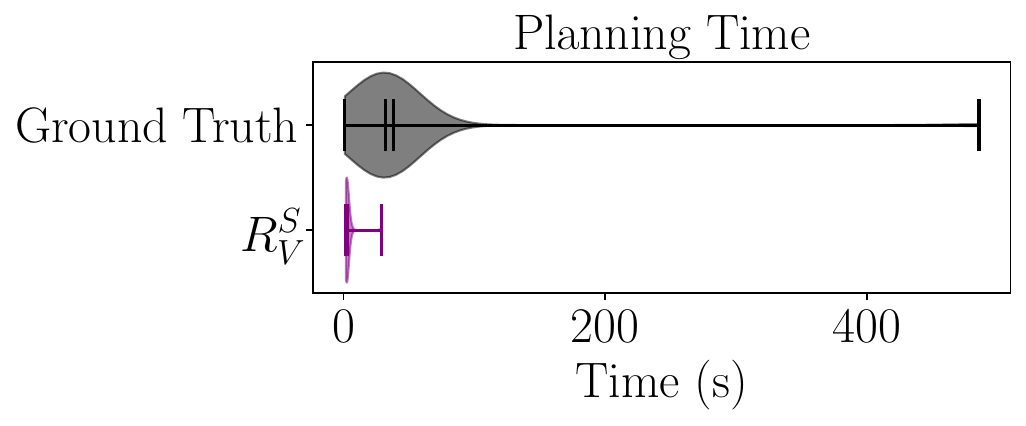}
    \caption{Distributions of planner time for \ragsearchcheck\ (including time for RAG) against a ground truth PDDL with the full state, for tasks successfully solved by \ragsearchcheck. \ragsearchcheck\ is far more efficient by only providing relevant context to the planner.}
    \label{graph:plan_times}
\end{figure}

Here are our findings:
\begin{enumerate}
    \item Based on Table~\ref{sim:results}, RAG reduces model hallucinations to significantly increase update accuracy. Without RAG, state updates succeed or fail inconsistently, as evidenced by \none\ having a higher success rate than \checkonly.
    \item Syntactically verifying LLM-generated state changes also improves state update accuracy (at least in ablations with RAG). As evidenced by the outliers in Figure \ref{graph:state_change_tokens}, few cases exist where re-prompting is invoked and extra tokens are consumed. KG context generally informs the LLM on proper KG triplet syntax, but verification serves as a stronger guarantee of correctness.
    \item Plan-generation is highly sensitive to knowledge graph accuracy. The sparsity of the instantiated environment allows the \static\ baseline to still achieve 67.5\% plan-generation accuracy. Even with 71.5\% state update accuracy, \none\ only marginally improves on plan-generation to 72.5\% accuracy. In most failure instances, the LLM-generated PPDL \texttt{:goal} is accurate, but the planner produces an inaccurate plan or fails to generate one due to an inaccurate KG state (which forms the PDDL problem \texttt{:init}). This leads to plans that (a) only partially solve the task or (b) cannot be executed due to inconsistency with the environment.
    \item L3M+P is not perfect: the usual cause of failure is when KG retrieval does not include all relevant context for a given NL update or plan, so the LLM produces hallucinated output from insufficient information.
    \item RAG enables a cost savings: As shown in Figure \ref{graph:state_change_tokens}, RAG uses fewer tokens to process successful state changes. On average, \ragsearchcheck\ uses 67.6\% fewer total input/output tokens compared to \none\ to process NL state changes (including tokens used to perform RAG).
    \item RAG enables a time savings: In Figure \ref{graph:plan_times}, \ragsearchcheck\ shows a speedup when comparing (a) combined RAG and planner time for \ragsearchcheck, against (b) only planner time with the full KG as the \texttt{:init} (ground truth). The average speedup is 12.5x for correctly-generated plans.

\begin{tcolorbox}[
    standard jigsaw,
    title=Example of a Successful State Change and Resulting Plan Produced by \ragsearchcheck,
    opacityback=0,
    fontupper=\small,
    fontlower=\small,
    breakable
]
\setlength{\parindent}{15pt}
\noindent
\textbf{State Change:} Kathleen placed the red pen on the 3rd level of the shelf in Jerry's bedroom.\\\\
\textbf{Prompt:} \textcolor{purple}{State Change} + \textcolor{purple}{Retrieved KG Context} + Determine which of the relations should be removed and what
new relations should be added...\\\\
\textbf{GPT-4o (generated KG update):}\\
\texttt{REMOVE: (red\_pen, in\_hand, kathleen)}\\
\texttt{ADD: (red\_pen, placed\_at\_shelf, jerry\_bedroom\_shelf), (red\_pen, on\_shelf\_level, shelf\_level\_3)}
\tcblower
\textbf{Task:} Place the red pen on the 5th level of the shelf in Alexander's bedroom.\\\\
\textbf{Retrieved KG context (\texttt{:init} block):}\\
\texttt{(red\_pen, placed\_at\_shelf, jerry\_bedroom\_shelf)}\\
\texttt{(red\_pen, on\_shelf\_level, shelf\_level\_3)}\texttt{\ldots}\\\\
\textbf{Prompt:} \textcolor{purple}{Task} + \textcolor{purple}{Retrieved KG Context} + Provide the goal block for a problem PDDL file... \\\\
\textbf{GPT-4o (generated \texttt{:goal} block):}\\
\texttt{(:goal (and (on\_shelf\_level bisque\_pen shelf\_level\_5) (placed\_at\_shelf bisque\_pen alexander\_bedroom\_shelf)))}\\\\
\textbf{Plan:}\\
\texttt{...}\\
\texttt{(place\_at\_shelf red\_pen alexander\_bedroom\_shelf the\_agent alexander\_bedroom shelf\_level\_5)}
\end{tcolorbox}

\end{enumerate}

\subsection{Embodied Interactive Simulation}

\label{sec:ai2_thor_exp}
We compare task completion rates of L3M+P that incorporate both updates versus relying solely on robot perception in a variety of home tasks in the embodied AI simulator AI2-THOR~\cite{Kolve2017AI2THORAn}. Scenes are selected from the ProcTHOR-10k dataset~\cite{deitke2022}. The output of a robot perception system is simulated by providing the semantic properties of the currently visible objects. Initial states and goal conditions are generated randomly for the following tasks: pick and place, wash dishes, clean bed, discard broken objects, and refrigerate food. State changes are scheduled randomly and can be categorized as either a reset (e.g. washed dishes getting dirty again), an addition (e.g. new dishes becoming dirty), or a relocation (e.g. dirty dishes are moved from their last known locations). For each condition, 20 task instances are generated. L3M+P re-plans when receiving verbal updates, and the visual-only variant re-plans when execution failures occur, indicating the environment is different from its knowledge at the last planning time. The results reported in Table~\ref{Tab:ai2_thor_results} show much higher success rates when L3M+P leverages verbal interactions. While visual-only performance could be improved by active perception behaviors to update knowledge, using verbal updates does not delay the task. Due to simulation uncertainties, some communicated updates do not align with the actual effects, leading to L3M+P failures.

\begin{table}
\centering
\begin{tblr}{
  width = \linewidth,
  colspec = {Q[200]Q[110]Q[110]Q[110]Q[110]Q[110]},
  cells = {c},
  cell{1}{1} = {r=2}{},
  cell{1}{2} = {c=5}{0.7\linewidth},
  vlines,
  hline{1,3-5} = {-}{},
  hline{2} = {2-7}{},
}
Success Rate~\% ($\uparrow$) & Task Type &  &  &  &  \\
 & Pick \& Place & Wash Dishes & Clean Bed & Discard Broken & Keep Cold \\
L3M+P & 65 & 85 & 90 & 70 & 65 \\
Visual Only & 15 & 0 & 5 & 10 & 55 
\end{tblr}
\caption{Task completion rates on the 5 types of tasks by using L3M+P versus the variant with only visual updates.}
\label{Tab:ai2_thor_results}
\end{table}

\subsection{Robot Demonstration}
\label{sec:robot_demo}
We present a robot demonstration in a scenario that builds upon the motivating example in Fig.~\ref{graph:motivating_example}. In the first task, Person A requests an apple. Initially, the robot plans to retrieve one from the fridge but instead picks up an apple spotted on the kitchen counter. Shortly after, Person B asks for a container to hold cereal. The robot has seen a mug on the table next to person A in the first task. However, before picking it up, Person A informs the robot that the mug is not empty. The robot then selects a bowl instead and retrieves the cereal from the previously scanned kitchen counter. Further details are available in \href{https://www.youtube.com/watch?v=Hs7GKJb55fA}{this supplementary video}.
\section{RELATED WORK}
\label{sec:related_work}

Several frameworks have been proposed for LLM-based planning and reasoning \cite{ahn2022do, huang2022language, yao2023react, song2023llmplanner, bhat2024grounding}. While these frameworks demonstrate LLM potential for plan generation, they may overestimate the extent to which language modeling translates to robust reasoning. Additionally, these works only focus on planning and do not attempt to design end-to-end solutions where the agent must both gather information about the environment and use that information to plan. 

Some recent frameworks address planning in dynamic environments. LLM-DP \cite{dagan2023llmdp} maintains a set of known information and a set of beliefs about the environment, both gradually updated as the agent makes observations. The beliefs are fed as context to an LLM to generate a PDDL world specification and goal for a classical planner. However, not only was this system was exclusively tested in Alfworld \cite{shridhar2021alfworldaligningtextembodied}, a text-based environment that only hosts a very specific problem domain, LLM-DP assumes that state changes are only caused by the agent and does not maintain knowledge of the environment across different planning tasks.

Statler \cite{yoneda2023statler} maintains and updates a world state used as a persistent memory for an LLM agent. The current state, stored in a JSON-like format, is used as context for planning tasks and is updated based on performed actions. Like LLM-DP, Statler is unable to account for external changes to the environment. Additionally, a specific world-state format must be curated for different domains. Even if the world-state representation were made domain-agnostic, the system would break down in complex environments with many objects and relations, increasing the likelihood of LLM hallucination.

Recent work has also explored memory storage for general-purpose LLM agents. A-MEM \cite{xu2025amemagenticmemoryllm} uses a flexible graph structure to organize information, populating nodes with atomic notes and assigned tags to dynamically insert inter-node links based on similarity. A-MEM is an orthogonal approach to L3M+P for using a graph to maintain an agent memory--while A-MEM uses a graph to loosely connect related information for more accurate retrieval, L3M+P uses a graph to explicitly represent relationships, and the benefits of accurate retrieval are a result of the graph structure.

Furthermore, Generative Agents \cite{park2023generativeagentsinteractivesimulacra} offers an approach to mimic human behavior with LLM agents in open-ended environments using dynamic memory, reflection, and planning. The agent records experiences over extended periods in a NL memory stream, and retrieves this information to generate context-aware actions. This work addresses a separate problem from L3M+P, which is to enable creative behavior in agents. As such, Generative Agents uses a loosely-defined memory, while L3M+P uses a structured and verified KG that is grounded in the real world to guarantee correctness.

\section{CONCLUSION}

We propose L3M+P, a framework that extends LLM and planning systems to support lifelong deployments of general-purpose service robots. L3M+P maintains a dynamic memory base that is continuously updated based on various, multi-modal sources of information about the environment, and remains in a consistent format that is queried for writing planning specifications. At planning time, relevant context from this memory is extracted to generate accurate specifications about the environment and task, which can then be fed to classical planners. Using an external memory base allows representing the world state of fairly complex environments in a domain-agnostic manner, and context retrieval prevents issues with limited context windows/hallucinations. In future work, the system could be enhanced to detect when re-planning is necessary in environments with frequent updates.


\section*{Acknowledgements}
This work has taken place in the Learning Agents Research
Group (LARG) at UT Austin.  LARG research is supported in part by NSF
(FAIN-2019844, NRT-2125858), ONR (W911NF-25-1-0065), ARO
(W911NF-23-2-0004), DARPA (Cooperative Agreement HR00112520004 on Ad
Hoc Teamwork) Lockheed Martin, and UT Austin's Good Systems grand
challenge.  Peter Stone serves as the Chief Scientist of Sony AI and
receives financial compensation for that role.  The terms of this
arrangement have been reviewed and approved by the University of Texas
at Austin in accordance with its policy on objectivity in research.

\bibliographystyle{IEEEtran}
\bibliography{references}


\end{document}